\pdfoutput=1

\documentclass[11pt]{article}

\usepackage{EMNLP2023}
\usepackage{times}
\usepackage{latexsym}
\usepackage{graphicx}
\usepackage{svg}
\usepackage{amsmath}
\usepackage{booktabs}

\usepackage[T1]{fontenc}

\usepackage[utf8]{inputenc}

\usepackage{microtype}

\usepackage{inconsolata}

\usepackage{soul}

\newcommand{\bs}[1]{{\textcolor{black}{#1}}}

\usepackage{multirow}
\usepackage{tabularx}
\usepackage{booktabs}
\usepackage{array}
\usepackage{tipa}
\newcommand\Tstrut{\rule{0pt}{2ex}}         

\title{\textit{Hi Guys} or \textit{Hi Folks}?
Benchmarking
Gender-Neutral Machine Translation with the GeNTE Corpus}

\author{Andrea Piergentili,\textsuperscript{1,2 $\ast$}  Beatrice Savoldi,\textsuperscript{2 $\ast$} Dennis Fucci,\textsuperscript{1,2} Matteo Negri\textsuperscript{2}, Luisa Bentivogli\textsuperscript{2}\\
  \textsuperscript{1}University of Trento\\
  \textsuperscript{2}Fondazione Bruno Kessler\\
  {\tt apiergentili,bsavoldi,dfucci,negri,bentivo\{@fbk.eu\}} \\}

\begin{document}
\maketitle
\begin{abstract}
\newcommand\blfootnote[1]{%
  \begingroup
  \renewcommand\thefootnote{}\footnote{#1}%
  \addtocounter{footnote}{-1}%
  \endgroup
}
\blfootnote{\textsuperscript{$\ast$}The authors contributed equally.}
Gender inequality is embedded in our 
communication practices and perpetuated in translation 
technologies. 
This becomes particularly apparent 
when translating into grammatical gender languages,
where
machine translation (MT) often
defaults to masculine and stereotypical representations by making undue binary gender assumptions. 
Our work addresses the  rising  demand for inclusive language by focusing head-on on gender-neutral translation from English to Italian.
We start from the essentials: proposing a 
dedicated 
benchmark
and
exploring 
automated evaluation methods.
%
%
%
%
%
%
%
First, we
introduce GeNTE, a natural, bilingual 
test set for gender-neutral translation, 
whose
creation 
was informed by a survey on the perception and use of 
neutral language.
Based 
on GeNTE,
we then overview 
existing reference-based evaluation approaches, highlight their limits, and propose a reference-free method more suitable to assess gender-neutral translation.



\end{abstract}

\section{Introduction}
\label{sec:introduction}

Societal gender asymmetries and inequalities are reflected and perpetuated through language \citep{stahlberg2007representation, menegatti2017gender}. Such awareness has grown also 
within the 
Natural Language Processing (NLP)
field 
\citep{blodgett-etal-2020-language}, where extensive research
has highlighted how several 
applications suffer from gender bias \citep{sun-etal-2019-mitigating, sheng-etal-2021-societal}. 
%
As also noted by MT users themselves \citep{forbes,
dev-etal-2021-harms}, among 
these applications are translation systems used at large scale, which 
pose the concrete risk of misrepresenting 
gender minorities by over-producing masculine forms, while reinforcing binary gendered expectations and stereotypes 
\citep{savoldi-etal-2021-gender, lardelli-gromann-2022-gender-fair}. 


To foster greater inclusivity 
and break free from the constraints of masculine/feminine language, 
neutral
strategies have emerged and are increasingly adopted in academia \citep{apa2020publication}, institutions \citep{hoglund2023gendering}, and industry alike \citep{microsoft}.
These strategies aim 
to
overcome marked forms that treat the masculine gender as the 
conceptually generic, default human prototype (e.g., \textit{humankind} vs. \textit{mankind}) \citep{silveira1980generic, bailey2022based}.  Thus, they challenge gender norms and embrace all gender identities by 
avoiding gendered terms 
when unnecessary
(e.g. \textit{chair} vs. \textit{chairman}/\textit{chairwoman}) \citep{hord2016bucking}.
%
%
%
%

English, being at the forefront of inclusive language changes and with its limited gendered grammar \citep{ackerman2019syntactic}, 
has faced fewer obstacles in adapting to 
neutral forms, which have 
already 
been modeled into monolingual 
generative
tasks \citep{sun2021they,vanmassenhove-etal-2021-neutral}. 
As recently underscored by 
\citet{amrhein-etal-2023-exploiting},
however, the resources and approaches made available for English are not portable to grammatical gender languages. 
Such need for dedicated efforts is exemplified in Italian, where neutral solutions must navigate the extensive encoding of masculine/feminine marking  (e.g. \textit{the doctors are qualified} $\rightarrow$ it: \textit{\textbf{i}/\textbf{le} dottor\textbf{i}/\textbf{esse}} \textit{sono qualificat\textbf{i}/\textbf{e}}) through synonymy or more complex rephrasing \citep{Euguideline} (e.g. $\rightarrow$ \textit{il personale medico \footnotesize{\texttt{[the medical staff]}}}). 
%
%
%
%
While indeed more challenging, pursuing 
inclusivity 
in Italian
is relevant exactly because sexist attitudes are more visible and impactful in grammatical gender languages \citep{wasserman2009que}.
%
%
%
Nonetheless, the implementation of neutral language in MT remains to date a basically uncharted territory, 
despite the desirability of neutral outputs  under several circumstances where gender is ambiguous or irrelevant.

In light of the above, by focusing on English$\rightarrow$Italian as an exemplary and 
representative translation pair and direction, we hereby lay the groundwork toward gender-neutral MT. Starting from 
a survey aimed to understand the
challenges of
neutral  translation 
in cross-lingual settings, we provide the necessary tools and resources  to foster research on the topic by 
estimating gender neutral translation in MT.
Hence, our main contributions are:
\textbf{(1)}
A study on the feasibility of neutral translation, by surveying
the potential   
trade-off among fluency, adequacy, and neutrality;
 \textbf{(2)} The creation of 
GeNTE,\footnote{
\textbf{Ge}nder-\textbf{N}eutral \textbf{T}ranslation \textbf{E}valuation. In Italian, \textit{gente} means
\textit{folks},
a term 
used for inclusive greetings in lieu 
of 
``\textit{guys}''.
}
%
the first natural, parallel corpus designed to test MT systems' ability to generate neutral translations;
\textbf{(3)} A comprehensive analysis of the (un)suitability of existing automatic metrics to evaluate 
neutral translation. As an inherent benchmark component, we 
indicate an alternative solution capable to better assess the task.
\smallskip

\bs{We make the GeNTE dataset freely available at \url{https://mt.fbk.eu/gente/} and 
release the evaluation code under Apache License 2.0 at 
\url{https://github.com/hlt-mt/fbk-NEUTR-evAL}}.

\section{Background}
\label{background}


Emerging research has 
highlighted the importance of reshaping 
gender in NLP technologies in a more inclusive 
manner  \citep{dev-etal-2021-harms},
also through the representation of non-binary identities and 
language  \citep{wagner-zarriess-2022-gender, lauscher-etal-2022-welcome,ovalle2023m}.
Foundational works in this area 
have included 
several applications, such as coreference resolution systems \citep{cao-daume-iii-2020-toward, brandl-etal-2022-conservative}, intra-lingual fair rewriters \citep{amrhein-etal-2023-exploiting}, and 
automatic classification of gender-neutral  text \citep{attanasio2021mimic}.


In MT, the research agenda has mainly focused on the improvement of masculine/feminine gender translation. 
Along this line, different mitigation methods have been devised to ensure that 
unambiguous gendered
referents (e.g. \textit{\textbf{he}/\textbf{she}} \textit{is a doctor})  are properly resolved in the target language \citep{costa-jussa-de-jorge-2020-fine, choubey2021improving, saunders-etal-2022-first}.
These methods are often tested on synthetic template-based datasets such as WinoMT \citep{stanovsky-etal-2019-evaluating} or SimpleGEN \citep{renduchintala-williams-2022-investigating}.
%
As also stressed by \citet{saunders2023gender}, however,
in realistic scenarios MT systems are 
also confronted with ambiguous input sentences that do not convey any gender distinction (e.g., en:  \textit{I called the \textbf{doctor}}).
%
%
%
Nonetheless, to date the resources and solutions 
envisioned for resolving such cases into grammatical gender languages 
like
Arabic \citep{alhafni-etal-2022-user}, Italian \citep{vanmassenhove-monti-2021-gender}, 
Spanish, or
French \citep{rarrick2023gate} 
entail offering 
two possible translation outputs,
still constrained to binary gender forms (e.g., it: \textit{Ho chiamato} \textit{\textbf{il} dott\textbf{ore}} \textsc{masc} vs. \textit{\textbf{la} dottor\textbf{essa}} \textsc{fem}).\footnote{Such double-outputs are currently offered for short, ambiguous queries also by 
Google Translate and Bing.}

As an exception within the current MT landscape, 
\citet{cho-etal-2019measuring} and \citet{ghosh2023chatgpt}
investigate the preservation of gender-ambiguous pronouns
for Korean/Bengali$\rightarrow$English. Since English can already boast the well-established neutral pronoun \textit{they}, their study does not face the additional challenges of preserving such unmarked vagueness into grammatical gender languages. 
Such challenges are exemplified 
by
\citet{saunders-etal-2020-neural},
who created
parallel test and fine-tuning data to
develop MT systems able to generate
non-binary translations 
for English$\rightarrow$German/Spanish. 
However,
their 
target sentences are artificial -- created by replacing gendered morphemes and articles with synthetic 
placeholders -- thus serving only as a proof-of-concept. 
%
%
%
To the best of our knowledge, \citet{piergentili2023gender}  are the first to advocate the use of target gender-neutral rephrasings and synonyms as a viable paradigm toward more inclusive MT when gender is unknown or simply irrelevant. 
Despite this call to action, no concrete steps have been taken yet to 
actually 
facilitate 
research in this
%
%
%
%
direction, 
not even
toward
suitable benchmarks to recognize 
the neutral forms occasionally generated by 
current systems
\citep{savoldi-etal-2022-morphosyntactic}.

In light of the above, the path toward gender-neutral translation in MT is 
bottlenecked by the lack of 
dedicated datasets and automated evaluations. 
Here, we fill this gap so to guide and allow research on this novel topic. 
To this aim, we start in \S\ref{sec:survey} by first ensuring that 
gender-neutral language can enable acceptable translations, not 
being
perceived as inappropriate or intrusive.


\section{Surveying Gender-Neutral Translation}
\label{sec:survey}

Neutralization is a form of linguistic gender inclusivity that relies on the retooling of established forms and grammar \citep{gabriel2018neutralising}. 
According to
the review of several gender-inclusive public guidelines 
by \citet{piergentili2023gender}, these can range from \textit{i)}
simple 
word changes, like
omissions
or 
article/noun
replacements with epicene alternatives (e.g. 
\textit{\textbf{il} maestr\textbf{o}} vs. \textit{l'insegnante}\footnote{en: ``the teacher''.}), to \textit{ii)}
more complex reformulations, which might 
involve
altering the sentence structure (e.g. \textit{\textbf{i} mie\textbf{i} collegh\textit{i} vs. \textit{le persone con cui lavoro}).\footnote{en: ``my colleagues'' vs. ``the people I work with''.}} As such, to ensure neutrality, these solutions might have an effect in terms of brevity or perceived fluency.


While widespread 
in monolingual, institutional contexts \citep{Euguideline}, the 
use of neutral forms in cross-lingual settings 
requires to 
weigh
additional 
non-negligible 
factors. 
First, translations are bounded to a source text, whose meaning 
must be 
properly  rendered in the target language. 
Thus, more creative reformulations 
might collide with this instrinc constraint. 
Also, 
it might be not always clear-cut when neutral translations ought to be performed. 
This
is the case of masculine generics 
in the source language (e.g. \textit{All \textit{\textbf{firemen}}}): while they do not neatly fall under the idea of ``ambiguous'' input, their propagation to a target language clashes
with the goal of inclusive MT itself. 

These issues stand unaddressed:
the study 
by \citet{lardelli2023genderfair} represents the only empirical investigation on the feasibility of gender-neutral translation, but it is concerned with the cognitive effort that its realization poses
to
post-editors. 
%
%
%
%
Therefore, to better understand the implications of gender-neutral translation for a wider range of 
stakeholders, we carried out a preliminary analysis on English{$\rightarrow$}Italian by surveying the opinions of potential MT end-users.

\begin{table}[t]
\setlength{\tabcolsep}{1.5pt}
\scriptsize
\begin{tabular}{lll|c|c|cc}
   &  & \textsc{Questionnaire Example sentences} & \textbf{Eq.} & \textbf{\textsc{NT}} & \textbf{\textsc{GT}} &  \\
   \hline
     &  & \textbf{tot. responses} & 36.5 & 42.5  & 21 &  \\
   \hline
\textsc{a.} &
  \begin{tabular}[c]{@{}l@{}}\\ \textbf{GT}\\ \textbf{NT}\end{tabular} &
  \begin{tabular}[c]{@{}l@{}}\textit{Some metals may be toxic
  to 
  \textbf{man}}.
  \\  Certi metalli possono essere tossici 
  per 
  \textbf{l'uomo}.
  \\ Certi metalli [...] 
  per \textbf{gli esseri umani} \textsubscript
{\texttt{{[}humans}{]}}.
\end{tabular} &
   \begin{tabular}[c]{@{}l@{}}39.6\\ \\ \end{tabular} &
  \begin{tabular}[c]{@{}l@{}} 50\\ \\ \end{tabular} &
  \begin{tabular}[c]{@{}l@{}}10.4\\ \\ \end{tabular} &
  \\[1.5mm]
 \cmidrule{2-6}
  
\textsc{b.} &
  \begin{tabular}[c]{@{}l@{}}\\ \textbf{GT}\\ \textbf{NT}\end{tabular} &
  \begin{tabular}[c]{@{}l@{}}\textit{Does \textbf{anyone} wish to speak against the proposal?}\\  \textbf{Qualcuno} desidera intervenire contro la proposta? 
  \\\textbf{Ci sono interventi}\textsubscript
{\texttt{{[}Are there speeches}{]}} contro [...]
\end{tabular} &
   \begin{tabular}[c]{@{}l@{}}54.7\\ \\ \end{tabular} &
  \begin{tabular}[c]{@{}l@{}}31.6\\ \\ \end{tabular} &
  \begin{tabular}[c]{@{}l@{}}13.7\\ \\ \end{tabular} &
   \\[1.5mm]
 \cmidrule{2-6}

 \textsc{c.} &
  \begin{tabular}[c]{@{}l@{}}\\ \textbf{GT}\\ \textbf{NT}\end{tabular} &
  \begin{tabular}[c]{@{}l@{}}  \textit{Indonesia is dealing with one million \textbf{refugees}...} 
  
  \\ \textit{L'Indonesia ha un milione di \textit{\textbf{profughi}...}} 
  \\ \textit{L'Indonesia ha un milione di \textbf{esuli}}\textsubscript
{\texttt{{[}exiles}{]}}...
\end{tabular} &
   \begin{tabular}[c]{@{}l@{}}40\\ \\ \end{tabular} &
  \begin{tabular}[c]{@{}l@{}}23.2\\ \\ \end{tabular} &
  \begin{tabular}[c]{@{}l@{}}36.8\\ \\ \end{tabular} &
   \\[1.5mm]
 \cmidrule{2-6}
\end{tabular}
\caption{Questionnaire example of English sentences with translation alternatives. 
For each example, participants responses --\textit{ GT and NT are equivalent, NT is preferrable, GT is preferrable} -- are shown (percentage). }
\label{table:questionnaire}
\end{table}

\noindent{\textbf{Questionnaire.} 
Our survey was structured into two main parts. 
In part \textit{\textbf{(i)}}, we indirectly assessed linguistic acceptability: given a source English sentence paired with both a gendered (GT) and a neutral (NT) translation,
we asked participants to indicate whether they had a 
preference or 
found 
them to be equivalent\footnote{i.e. 
equally adequate and fluent.}
(see Table \ref{table:questionnaire}).
Then, 
in part \textbf{\textit{(ii)}} we asked
direct questions to gauge participants' use and attitude toward gender-neutral language. The questionnaire was distributed online and received 98  
responses by eligible participants.
While all
details  
are provided in Appendix \ref{app:questionnaire}, here 
we summarize  our main insights.


 \begin{figure}[t]
  \centering
  \includegraphics[scale=0.55]{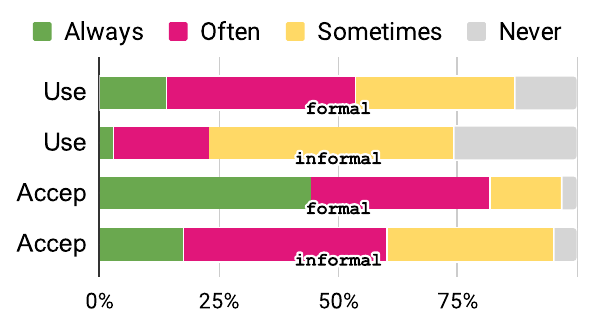} 
  \caption{ Frequency of use and 
  acceptance
  of   neutral language in \textit{formal} vs. \textit{informal} communication.
  }
  \label{fig:use-att}
\end{figure}

First, the linguistic acceptability of gender-neutral translations 
was positively judged, perhaps at the higher rate than our own expectations.
In fact, overall results indicate that in the majority of cases the NT was deemed preferable (42.5\%) rather than equivalent to the GT (36.5\%), where only a minority 
favoured gendered translations (21\%).
Possibly, such trend is explained 
in light of 
participants' ideological preference for inclusive language rather than by purely linguistic factors.  As shown in Table \ref{table:questionnaire}, this seems to be confirmed by disaggregating responses for each translated example sentence, where we can extract more qualitative insights.
%
Indeed, example \textsc{A} attests one of the strongest preferences for NT, signalling a negative attitude toward the 
propagation of default, masculine forms in GT.
%
Then, concerning the use of more of complex neutral rephrasings
(example \textsc{B}), we found that slightly longer and sentence-altering neutralization strategies were 
still 
considered largely acceptable. 
%
Instead, literal NT with limited changes, which however sacrificed more the 
source meaning  or altered its tone (example \textsc{c}),  were comparatively penalized.
%
%
%
This
trend was also confirmed 
in part \textit{(ii)}
of the survey  
(see Appendix \ref{app:questionnaire}).}
%
%
%
%
%
%
Finally, as shown in Figure \ref{fig:use-att}, participants' responses to direct questions attest that neutralization strategies are accepted/used differently depending on the speech situation, with a  preference for their use in formal communicative situations.
%
Having confirmed the feasibility and overall acceptability
of neutral translations, we embed the gathered qualitative considerations in the design of the GeNTE corpus (\S\ref{sec:corpus}). 

\section{The GeNTE corpus}
\label{sec:corpus}


%
GeNTE
is the first
test set designed to evaluate MT models' ability to perform gender-neutral translations, but only under desirable circumstances.
In fact, when referents' gender is unknown or irrelevant, undue gender inferences 
should not
be made and translation should be neutral. 
However, neutralization should not be always enforced; for instance, when a referent's gender is relevant and known, MT should not over-generalize to neutral translations.
%
The corpus hence 
consists of 1,500 English-Italian parallel sentences with mentions to human referents that equally represent two translation scenarios:  
\textbf{\textbf{1)}} \texttt{Set-N}, featuring 
gender-ambiguous source 
sentences that require to be neutrally rendered 
in translation; 
\textbf{\textbf{2)}} \texttt{Set-G}, featuring 
gender-unambiguous source sentences, which shall 
be properly rendered with gendered (masculine or feminine) forms in translation.
%
%
%
%
Altogether, these sets allow to benchmark whether systems are able to perform gender-neutral translation, and if they do so  when appropriate. 


%
We build GeNTE on naturally occurring instances  of both 
scenarios retrieved 
from 
Europarl \citep{koehn2005europarl}. 
Besides
being a widely popular and high-quality MT resource, we chose this 
corpus
inasmuch it represents formal communicative situations from the administrative/institutional domain. Accordingly, it reflects the context for which gender-neutral forms are traditionally intended, 
also in line with the stakeholders' preference highlighted in
\S\ref{sec:survey}.
Also, as examined by \citet{saunders-2022-domain}, Europarl exhibits a large amount of gender-ambiguous cases that -- although translated with gendered forms in the original references of the corpus --
lend themselves as suitable candidates for  
neutralization.  
As explained in the forthcoming paragraphs (\S 
\ref{sec:references}), for each of these original Europarl gendered target sentences, we create an additional gender-neutral reference translation.

\subsection{Data selection and annotation}
\label{selection}

\paragraph{Data extraction.}
To retrieve Europarl\footnote{
\url{https://www.statmt.org/europarl/archives.html}} segments representing our two translation scenarios of interest, 
we 
crafted
regular expressions
to: \textit{i)} identify source sentences containing mentions to human referents, \textit{ii)} maximize the variability of linguistic phenomena included in the corpus, and \textit{iii)} ensure a balanced distribution of both unambiguous and ambiguous gender translation cases. 
To this 
aim, we 
targeted \texttt{Set-G} segments by matching source English sentences that contained explicit gender cues, e.g. lexically gendered words (\textit{sister}, \textit{woman}),  titles (\textit{Mr}, \textit{Mrs}) and marked pronouns (\textit{him}, \textit{her}). 
\texttt{Set-N}, instead, was populated by matching several word classes that do not convey any gender distinction in English (e.g. \textit{you}, \textit{citizens}, \textit{went}), but 
typically 
correspond to masculine/feminine expressions in the target language. Also,
we searched for
masculine terms used generically, such as \textit{man} and its 
derived 
compounds (e.g., \textit{chairman}, \textit{layman}). 
In fact, masculine generics are  unreliable gender cues and, following the survey findings (\S\ref{sec:survey}), should not be propagated in MT. 


\paragraph{Sentence editing.}
On the collected material, a first 
intervention was carried out 
to streamline the evaluation of gender-neutral translation. In fact, some of the source 
sentences contained mentions of 
multiple
referents, 
which required
the combination of different 
forms in 
translation 
(i.e. neut/masc/fem).
In those cases, the parallel sentences
were manually edited so as to ensure that they only 
include  referents that require
the same type of (either neutral or gendered) forms. 
In this way, each sentence 
pair 
can be handled as a whole coherent unit, thus avoiding the complexities of evaluating intricate combinations of phenomena. 
To ensure a balanced distribution of  instances from both \texttt{Set-N} and \texttt{Set-G}, a second intervention was required to compensate for the under-representation of unambiguous cases.\footnote{Such lack confirms the vast representation of generic and unknown referents in Europarl, as found in \citet{saunders-2022-domain}.}
Although these edits slightly 
reduce
the naturalness of the data, they allow for a simpler and 
sound evaluation, crucial 
to shed light on a
complex task such as gender-neutral MT.
Instead, other edits were made to enhance the quality of the corpus; all of them are reported in Appendix \S\ref{app:data_editing}. Once the editing phase was concluded, all sentence pairs were annotated as N in \texttt{Set-N}, and as F or M in \texttt{Set-G}. 
In the annotation process, it was verified that the initial pool of -- automatically extracted -- candidate sentences were correctly 
assigned to
\texttt{Set-N} and \texttt{Set-G} by accounting for the sentence context. In this way, we could differentiate between the use of gendered words as either masculine generics (e.g. \textit{It is up to an accused employer to prove \textbf{his} innocence} -- identified as N) or as informative of a referent’s gender (e.g. \textit{I would like to thank Commissioner Byrne for \textbf{his} cooperation.} -- identified as G).


\begin{table*}[htp!]
\scriptsize{
    \centering
    \begin{tabularx}{\textwidth}{clX}
    \hline
    \multirow{5}{*}{\textit{i} -- \textsc{N}} & \textbf{SRC} & I, along with \textbf{all my colleagues}, wish to welcome this [...] \Tstrut \\ 
    & \textbf{REF-G} & Insieme a \textbf{tutti i miei colleghi}, desidero esprimere il mio compiacimento per questa [...]  \\ 
    & \textbf{REF-N 1} & Insieme \textbf{agli altri membri}\textsubscript
{\texttt{{[}other members}{]}}, desidero esprimere il mio compiacimento per questa [...]  \\ 
    & \textbf{REF-N 2} & Insieme a \textbf{ogni collega}\textsubscript
{\texttt{{[}each colleague}{]}}, desidero esprimere il mio compiacimento per questa [...]  \\ 
    & \textbf{REF-N 3} & Insieme a \textbf{tutte le persone con cui lavoro}\textsubscript
{\texttt{{[}all the persons with whom I work}{]}}, desidero esprimere il mio compiacimento per questa [...] \\ \hline
    \multirow{5}{*}{\textit{ii} -- \textsc{M}} & \textbf{SRC} & I welcome this excellent report \textbf{from my colleague} \underline{Mr} Skinner. \Tstrut  \\ 
    & \textbf{REF-G} & Valuto positivamente la relazione \textbf{del collega}, onorevole Skinner.  \\ 
    & \textbf{REF-N 1} & Valuto positivamente la relazione \textbf{dell'onorevole collega}\textsubscript
{\texttt{{[}of the honorable colleague}{]}} Skinner.  \\ 
    & \textbf{REF-N 2} & Valuto positivamente la relazione \textbf{dell'onorevole collega} Skinner.  \\ 
    & \textbf{REF-N 3} & Valuto positivamente la relazione \textbf{dell'onorevole collega} Skinner.  \\ \hline
    \multirow{6}{*}{\textit{iii} -- \textsc{F}} & \textbf{SRC} & \underline{Mrs} Ana de Palacio Vallelersundi has \textbf{a \underline{sister} who is a Commissioner} [...] \Tstrut \\ 
    & \textbf{REF-G} & \textbf{La} onorevole [...] ha \textbf{una sorella, la quale è una Comissaria} [...]  \\ 
    & \textbf{REF-N 1} & N.A. \\
    & \textbf{REF-N 2} & L'onorevole [...] ha \textbf{uno stretto legame di parentela}\textsubscript
{\texttt{{[}is closely related}{]}} con \textbf{un membro della Commissione}\textsubscript
{\texttt{{[}a member of the Commission}{]}} \\ 
    & \textbf{REF-N 3} & N.A.  \\\hline  
    \end{tabularx}
\caption{Examples of entries in the \textsc{common-set}. REF-G indicates the gendered references, REF-N 1, 2, 3 indicate the neutralized references produced by Translator 1, 2, and 3 respectively. Words in \textbf{bold} are mentions of human referents; \underline{underlined}  words are linguistic cues informing about the referents's gender.}
\label{tab:entries}
 }
\end{table*}

\subsection{Creation of gender-neutral references}
\label{sec:references}

As a confirmation of the predominant use of gendered forms 
when translating into
grammatical gender languages, 
it is worth remarking that almost all (97.2\%) segments collected from Europarl have gendered references in Italian.
Inspired by the design of natural (binary) gender bias benchmarks such as MuST-SHE \citep{bentivogli-2020-gender} and MT-GenEval \citep{currey-etal-2022-mt}, we thus created a second translation, so to allow for a reference-based contrastive evaluation of gender-neutral MT  (see $\S$\ref{sec:reference-based}).
%
%
To this aim, for each sentence pair, we created an additional Italian reference, which differs from the original one only in that it refers to the human entities with neutral expressions.
This makes it possible to isolate gender-related linguistic elements as the only source of variation in the score of system outputs when evaluated against both the gendered and the neutral references.
As 
neutralization is an open-ended task that entails a high degree of variability in the possible solutions, we wanted such variability 
accounted for in the neutral references.  Therefore, their creation was assigned to three professional translators hired 
via a translation agency.\footnote{The cost paid to the agency was of 60 euros/hour, for a total of 14 hours of work for each translator.} 
Each of them was assigned a different portion of the collected Italian references, to be post-edited so as to \bs{only replace 
gendered terms with neutral formulations.}
An  expert linguist native speaker of Italian\footnote{The linguist is one of the authors of the paper.} prepared 
 detailed 
instructions
\footnote{Released together with the GeNTE corpus.}
drawing from existing 
guidelines 
for the institutional domain. 
After an initial training session, the linguist 
supported
the translators 
throughout the process 
and finally checked all the neutralizations.
In Appendix \ref{app:ref_creation}, we provide  qualitative insights regarding revisions and supervision of the linguist.


\begin{table*}[h]
    \centering
\scriptsize{
    \begin{tabular}{c|c|c||c|c||c|c||c}
        \multicolumn{8}{c}{\textbf{\textsc{GeNTE}}} \Tstrut \\ \hline
        \multirow{2}{*}{} & \multicolumn{2}{c||}{\textbf{Source}} & \multicolumn{2}{c||}{\textbf{REF-G}}  & \multicolumn{2}{c||}{\textbf{REF-N}} & \Tstrut \\ \cline{2-8} 
        ~ & \# sentences & Avg length 
        & \# sentences & Avg length 
        & \# sentences & Avg length & \# G-words
        \Tstrut\\ \hline
        \textbf{\texttt{Set-N}} & 750 & 25.67 
        & 750 & 24.66 
        & 750 & 26.95  & 1,972
        \Tstrut \\ 
        \textbf{\texttt{Set-G}} & 750  & 26.51 
        & 750 & 25.26 
        & 750 & 26.55 & 2,148

        \\ \hline 
        \multicolumn{8}{c}{\textbf{\textsc{COMMON-SET}}} \Tstrut  \\ \hline
         & \multicolumn{2}{c||}{\textbf{Source}} & \multicolumn{2}{c||}{\textbf{REF-G}}  & \multicolumn{2}{c||}{\textbf{REF-N}} & \Tstrut \\ \cline{2-8}
        ~ & \# sentences & Avg length & 
        \# sentences & Avg length 
        & \# sentences & Avg length & \# G-words
        \Tstrut \\ \hline
        \textbf{\texttt{Set-N}} & 100 & 27.45 
        & 100 & 27.00 
        & 300 & 28.87 & 300
        \Tstrut \\ 
        \textbf{\texttt{Set-G}} & 100  & 26.99 
        & 100  & 26.34 
        & 300  & 27.57 & 299
        \\ \hline
    \end{tabular}
    }

\caption{Corpus statistics for \textsc{GeNTE} and its subset \textsc{common-set}. Both sets requiring gendered translations (\texttt{Set-G}) are equally balanced between \textsc{F} and \textsc{M} sentences. Average lengths are calculated ignoring punctuation. In the last column, we provide the number of gendered words in the REF-Gs that had to be neutralized in the REF-Ns.}  
\label{tab:corpus_stats}

\end{table*}

\paragraph{GeNTE~\textsc{common-set.}} 
%
%
%

Whereas each translator was in charge of post-editing one given portion of the corpus, we also selected a common set of 200 references to be neutralized by all translators (henceforth referred to as the \textsc{common-set}); 100 were taken from the gendered set (\textsc{common-set-g}), and 100  from the neutral one (\textsc{common-set-n}). Thus, we obtained 200 source sentences,
each paired with one (original) gendered reference and three (post-edited) neutral references. 
The creation of a \textsc{common-set} was primarily motivated by the goal of having a subset of the corpus that could be used to test the robustness of evaluation protocols and metrics  
across the three different neutral references (see \S\ref{sec:reference-based}). Orthogonally, it 
allowed us to measure linguistic variability among the neutral and gendered references (see Appendix \ref{sec:ling_div}).
Table \ref{tab:entries} 
shows examples 
from the \textsc{common-set}, which confirm the 
findings of our preliminary survey (\S\ref{sec:survey}). Example \textit{i} is representative of the variability that is inherent to the 
neutralization task. Example \textit{ii}, instead shows 
a
rare situation where all 
translators used the 
same neutralization device and produced an identical sentence. Finally,  \textit{iii} shows 
a 
gendered term, whose neutralization requires verbose periphrases that compromise
the original text's fluency and
style. 
This case was signaled as particularly difficult to neutralize:
two translators out of three did not create a neutral reference.
Overall, based on a manual analysis of the \textsc{common-set}, the translators produced three identical gender-neutral references in 13.57\% of the cases, while an additional 8\% 
of translations 
exhibited a high degree of similarity (e.g., the same neutral words are used, but in a different order). These statistics are positive: they show that the GeNTE \textsc{common-set} exhibits a good level of variability ($\sim$79\%), which is desirable to test open-ended generation tasks like MT. Also – and especially in light of the fact that the translators worked independently – the $\sim$21\% of identical/similar neutralizations suggests that neutralizing translation is a 
challenging but 
feasible task.

To conclude, 
relevant statistics for GeNTE and its \textsc{common-set}
are provided in  Table \ref{tab:corpus_stats}.

\section{Gender-neutral Evaluation Protocols}
\label{sec:evaluation}


We complement our benchmark creation effort with a study on the possible approaches for using GeNTE to conduct automated evaluations of neutral MT. 
To this aim, we  first define sound test-bed conditions
(\S\ref{sec:test-bed}). On this basis, we  then experiment  with a contrastive, reference-based protocol  to inspect the effectiveness of standard MT  metrics to assess neutral translation
(\S\ref{sec:reference-based}). Then, to overcome the limitations encountered with the reference-based approach, we implement a reference-free protocol (\S\ref{sec:reference-free}), which shows promise in advancing the task's evaluation.

\subsection{Test-bed}
\label{sec:test-bed}

To ensure a sound comparison between different automatic evaluation protocols, we built a test-bed based on  the GeNTE \textsc{common-set} (\S\ref{sec:references}, Table \ref{tab:corpus_stats}). Our test-bed includes  relevant instances in relation to our task, namely gendered and gender-neutral automatic translations in equal proportion. On this basis, the analyzed 
evaluation approaches
can
be compared in their 
ability to reward systems that generate neutralized outputs only when due.

The automatic Italian translations of the \textsc{common-set} 
sources
were generated with two leading  commercial MT systems: 
Amazon Translate\footnote{\url{https://aws.amazon.com/translate/}} and 
DeepL.\footnote{\url{https://www.deepl.com/en/translator}}
However, a manual inspection showed an almost complete lack of representation of gender-neutral
%
translations in the outputs:  gendered translations were generated
for all but one
of the \textsc{common-set-n} inputs.\footnote{This provides a glimpse into the shortcomings of inclusivity within the current MT landscape.}
This result revealed the unsuitability of such outputs for investigating the automated evaluation of neutral translation itself.
Accordingly, to 
obtain
neutral 
(MT-like)
outputs to be included in the test-bed, we resorted to 
manually post-editing
the 100 \textsc{common-set-n} 
translations generated with undue gender assignments.
To do so, we leveraged our manually-created neutral references ($\S$\ref{sec:references}): we substituted the neutral forms produced by the three professional translators to the gendered forms in the MT outputs,  
%
%
so as to make them neutral without altering the rest of the sentence.\footnote{On average, 12\% of the words present in the systems’ output were substituted through post-editing, thus these edits have a minimal and circumscribed impact that does not alter the original output sentence.
}
For each 
system,
we thus obtained 
three
sets of neutral output sentences (one per translator), 
so to account for the robustness of different evaluation methods to the linguistic variability expressed in the 
inventory of neutralization strategies potentially applicable by humans and machines.
%



\begin{table*}[t]
\centering
\scriptsize{
    \begin{tabular}{l||r|r|r|r|r|r||r|r|r|r|r|r}
    \hline
        \multirow{3}{*}{\textbf{Metric}} & \multicolumn{6}{c||}{\textsc{\textbf{common-set-g}}}  & \multicolumn{6}{c}{\textsc{\textbf{common-set-n}}}  \\ \cline{2-13}
        ~ & \multicolumn{3}{c|}{\textbf{DeepL}} & \multicolumn{3}{c||}{\textbf{Amazon}} & \multicolumn{3}{c|}{\textbf{DeepL}} & \multicolumn{3}{c}{\textbf{Amazon}} \Tstrut \\ \cline{2-13}
        ~ & \textbf{REF-G} & \textbf{REF-N} & \textbf{$\Delta$\%} & \textbf{REF-G} & \textbf{REF-N} & \textbf{$\Delta$\%} & \textbf{REF-N} & \textbf{REF-G} & \textbf{$\Delta$\%} & \textbf{REF-N} & \textbf{REF-G} & \textbf{$\Delta$\%} \Tstrut \\ \hline 
        \textbf{BLEU} & 34.95 & 25.76 & \textbf{26.30} & 35.20 & 25.75 & \textbf{26.83} & 24.86 & 23.78 & \textbf{4.35} & 24.38 & 22.83 & \textbf{6.36} \\
        \textbf{chrF} & 64.18 & 56.87 & \textbf{11.40} & 64.01 & 56.67 & \textbf{11.46} & 55.47 & 56.03 & -1.00 & 55.52 & 55.70 & -0.32 \\
        \textbf{TER} $\downarrow$ & 52.18 & 61.92 & \textbf{18.66} & 53.54 & 63.82 & \textbf{19.21} & 66.50 & 69.87 & \textbf{5.06} & 66.68 & 70.96 & \textbf{6.42} \\
        \textbf{METEOR} & 62.10 & 51.63 & \textbf{16.87} & 60.90 & 50.40 & \textbf{17.24} & 48.30 & 47.99 & \textbf{0.65} & 47.73 & 47.30 & \textbf{0.89} \\
        \textbf{BERTscore} & 88.34 & 85.51 & \textbf{3.20} & 88.00 & 85.18 & \textbf{3.20} & 84.54 & 84.57 & -0.03 & 84.37 & 84.39 & -0.02 \\
        \textbf{COMET} & 87.89 & 85.40 &\textbf{2.84} & 87.36 & 85.11 & \textbf{2.57} & 84.97 & 85.26 & -0.34 & 84.75 & 85.06 & -0.37 \\
        \textbf{BLEURT} & 80.50 & 76.10 & \textbf{5.47} & 79.67 & 75.92 & \textbf{4.71} & 76.33 & 76.58 & -0.33 & 75.35 & 75.71 & -0.48 \\ \hline
    \end{tabular}
    }
%
\caption{
%
%
Corpus-level scores for DeepL and Amazon Translate, 
and percentage \textbf{gains} ($\Delta$\%, with sign changed for TER) with respect to the correct references. 
\textsc{common-set-g}: the original MT output is evaluated against each of the three available references, resulting scores are averaged. 
\textsc{common-set-n}:
each of the three edited MT outputs is evaluated against the two references not used to neutralize it, all resulting scores are averaged. 
}
\label{tab:results_corpus_level}
\end{table*}

\subsection{Reference-based Evaluation}
\label{sec:reference-based}

\subsubsection{Setting}



In this evaluation protocol we aim to verify whether common reference-based MT metrics can be effectively used to identify gendered and neutral translations.
The protocol is based on the idea that 
if a system generates a gendered translation, its output will be rewarded when evaluated against a gendered reference and penalized when evaluated against a neutral one.
On the contrary, if a system produces a neutral translation, this is expected to be rewarded when compared to a neutral reference and penalized when compared to a gendered one.





\paragraph{Contrastive Protocol.}

Given a system output and a reference-based metric, we compute corpus-level scores against both the
gendered and the neutral references provided in \textsc{common-set}.
Then, for \textsc{common-set-g} the metric is effective
if the scores are higher when computed against the gendered translations than the neutral ones; vice versa for \textsc{common-set-n}.

\paragraph{Metrics.}
We study the 
effectiveness
of 
a set of widely used metrics.
These can be categorized as 
\textit{i)} n-gram 
overlap
metrics: BLEU\footnote{\texttt{\bs{BLEU|\#:1|c:mixed|e:no|tok:13a|s:exp|v:2.3.1}}}
\citep{papineni-etal-2002-bleu}, chrF \citep{popovic-2015-chrf}, TER \citep{snover-ter-2006}, and METEOR \citep{banerjee-lavie-2005-meteor}, 
which are sensitive to 
surface form differences between outputs and references \cite{glushkova2023bleu};
\textit{ii)} neural model-based metrics: BERTScore \citep{zhang2020bertscore}, BLEURT \citep{sellam-etal-2020-bleurt}, and COMET \citep{rei-etal-2020-comet},
which compare 
semantic representations based on the respective 
underlying 
models.

 \subsubsection{Results}
 \label{ref-based-results}
Table \ref{tab:results_corpus_level} reports the results 
computed with each metric on our test-bed.
First, results are consistent between 
DeepL and Amazon Translate. 
With respect to \textsc{common-set-g}, all the metrics correctly 
give higher scores 
for
gendered references than 
for
neutral ones:
%
positive percentage differences thus indicate that the metrics correctly reward systems' gendered translations.
%
%
%
However, in \textsc{common-set-n} 
there is a divergence in performance between 
n-gram overlap
metrics and neural metrics.
Only three
metrics based on n-gram overlap -- BLEU, TER, and METEOR -- correctly assign higher scores to systems evaluated against the neutral references.

%
These results show that, compared to neural metrics, 
n-gram overlap
metrics appear more suitable for 
assessing
gender-neutrality.
%
The lower effectiveness of the neural metrics could be attributed to the lower frequency of neutral expressions in the training data of these models, leading to a lower probabilities assignment.
Also, neural metrics
are sensitive to semantic variations, but  
robust to surface lexical or morphological differences. 
%
Thus, since 
gender neutralizations 
preserve the essential semantics of their gendered equivalents, neural metrics are unable to properly frame their differences in this contrastive
setting.
This is evident in 
the consistently higher $\Delta$ percentages observed in \textsc{common-set-g}
for n-gram overlap metrics (all above 11\%) compared to the lower percentages obtained with neural metrics (all below 6\%).


We further experimented with BLEU, TER, and METEOR to investigate their ability to provide more fine-grained
evaluations. We thus tested them on the same data and the same contrastive principle, but at the sentence level (i.e. a neutral output sentence should obtain higher scores on the neutral reference compared to the gendered reference, and vice versa for a gendered output translation). 
In this way,  the $\Delta$ obtained with the contrastive reference-based evaluation protocol that relies on BLEU, TER and METEOR can be used to categorize each sentence as either gendered or gender-neutral: 
higher BLEU on the neutral reference wrt. the gendered reference $\rightarrow$ output sentence classified as neutral; higher BLEU on the gendered reference wrt. the neutral reference $\rightarrow$ output sentence classified as gendered.
By distinguishing the sentences belonging to \textsc{common-set-n} and \textsc{common-set-g} in advance, we can \bs{thus} calculate an overall accuracy that represents the proportion of sentences correctly categorized.
The results are presented in Table \ref{tab:results_acc}.
%
%
For \textsc{common-set-g}, the performance is rather promising for all the three metrics, with accuracy scores always 
above 90\% for the outputs of both Amazon Translate and DeepL. 
This is in line with the corpus-level results for the same set. Interestingly, however, we see that for \textsc{common-set-n} the accuracy scores are very low, close to 
random choice for BLEU and METEOR.
Only TER-based evaluations are higher (64.83\% for Amazon Translate, 66\% for DeepL).

In conclusion, through this closer inspection, we 
find that
%
%
%
none of the n-gram overlap metrics is 
actually reliable 
for the evaluation of gender-neutral translation.
%
%
This is possibly due to the fact that
our reference-based evaluation approach, just like the metrics it is based on, 
is heavily dependent on the reference sentences.
Outputs that deviate from the reference, even if they are acceptable translations, may therefore be penalized. 
This issue becomes particularly critical when evaluating gender-neutral translations, as periphrasis or synonymy are among the most common and accepted techniques used for achieving neutrality (§\ref{sec:survey}). 
These strategies are inherently penalized by n-gram overlap metrics and do not seem to entail significant differences in meaning according to neural metrics, 
thus advocating for alternative, reference-free solutions.

\subsection{Reference-free Evaluation}
\label{sec:reference-free}

\subsubsection{Setting}

%
%
%
%
%
Our reference-free protocol for the evaluation of gender-neutral translation explores a classification-based approach.
We cast the problem as a binary 
task to classify if 
automatically-translated sentences are 
gendered or neutral.
%
%
%
%
%
%
%
%
%
%
Implementing this procedure 
requires
\textit{i)} 
generating
training data, 
and \textit{ii)} 
training the classifier on the collected data.
Then, results are 
computed on our test-bed in terms of classification accuracy.

\paragraph{Synthetic Data Generation.}
Confronted with the lack of 
Italian corpora featuring gender-neutral language, we 
resorted
to synthetic data generation by prompting GPT (gpt-3.5-turbo). 
To do so, we devised a three-step approach that allowed for a more controlled generation procedure with reduced risk of noise (for full details, see Appendix \ref{app:synthetic}). 
First, similarly to \citet{attanasio2021mimic}, we manually created 800 triplets of 
neutral, masculine, and feminine
referents (e.g. the neighbours: \textit{ \textit{il vicinato} - \textbf{i} vicin\textbf{i}} - \textit{l\textbf{e} vicin\textbf{e}}). 
Then,
we used such triplets 
as
seedwords to prompt GPT and generate 
triplets of sentences, which only 
differ
for the inserted 
(neut/masc/fem)
seedword.
This resulted in 
$\sim$60,000
sentences with a very low level of noise, 
but featuring a 
rather
simple and repetitive syntactic structure. 
Therefore, we finally
carried out a second generation round, 
prompting GPT to rewrite each triplet adding context to increase sentence variability and length.
This led to a
final synthetic corpus of $\sim$380,000 sentences, equally distributed across neut/masc/fem instances, and with varied structures to 
favor generalization.\footnote{The synthetic corpus is released with the evaluation model.} 

%

\begin{table}[tb]
\centering
\small

\setlength{\tabcolsep}{2.9pt}
    \begin{tabular}{l|l|l|l||l|l|l}
    \hline
        \multirow{2}{*}{\textbf{Metric}} & \multicolumn{3}{c||}{\textbf{DeepL}} & \multicolumn{3}{c}{\textbf{Amazon}} \\ \cline{2-7}
         & \texttt{Set-G}  & \texttt{Set-N}  & All  & \texttt{Set-G}  & \texttt{Set-N}  & All \Tstrut \\ \hline
        \textbf{BLEU }& 92.00 & 51.83 & 71.92 & 93.00 & 52.66 & 73.00 \Tstrut \\
        \textbf{TER} & 90.50 & 66.00 & 78.25 & 90.50 & 64.83 & 77.67 \Tstrut \\ 
        \textbf{METEOR} & \textbf{94.50} & 48.67 & 71.59 & \textbf{94.50} & 47.33 & 70.92 \Tstrut \\ \hline
        \textbf{Classifier} & 91.00 & \textbf{88.67} & \textbf{89.83} & 87.00 & \textbf{87.33} & \textbf{87.17} \Tstrut \\ \hline
    \end{tabular}
\caption{Accuracy scores for reference-based (BLEU, TER, and METEOR) and reference-free (classifier) evaluation protocols.}

\label{tab:results_acc}
\end{table}

\paragraph{Gender-Neutral Classifier.}
To implement the classification model, 
we leveraged UmBERTo\footnote{\url{https://huggingface.co/Musixmatch/umberto-commoncrawl-cased-v1}}, 
a Roberta-based language model (LM) fine-tuned on the Italian section of the web corpus OSCAR \cite{ortizsuarez:hal-02148693}.
In the survey by \citet{tamburini-bertology}, UmBERTo
was proven to be one of the best-performing LMs for Italian.
Given a sequence of tokens, 
UmBERTo returns a contextualized vector for each token, 
including the special [CLS] token 
placed at
the beginning of the sentence. 
As suggested by \citet{devlin-etal-2019-bert}, 
we added a linear layer on top of the [CLS] vector to predict the neutral or gendered  class.
We trained the parameters of both the linear layer and UmBERTo on the classification task using the synthetic corpus labeled with neutral or gendered -- for feminine and masculine -- tags. Since this solution yielded the best results, our final classifier was trained in unbalanced data conditions, by making use of all synthetic gendered sentences (e.g. 1/3 fem and 1/3 masc) and all neutral sentences (1/3 neut). For complete details on the training setup see Appendix \ref{app:training_setup}.


\subsubsection{Results}
Compared to the 
accuracy scores 
obtained 
via
sentence-level contrastive evaluation based on BLEU, TER, and METEOR
(first three rows of Table \ref{tab:results_acc}), it is 
evident that the scores achieved by the trained classifier (row 4) are notably higher.
These discrepancies primarily arise from the performance on the neutral outputs (\texttt{Set-N}), where the classifier outperforms the 
best
n-gram overlap metric (TER) by a margin of up to 22.67 points for 
DeepL.
However, for gendered outputs, the classifier demonstrates slightly lower results compared to 
the three reference-based metrics
(except 
for
DeepL, where it outperforms TER).
As a result, 
for our reference-free approach,
the gap between the scores obtained on gendered and neutral outputs is small 
(especially for Amazon Translate), attesting a balanced performance 
across the two classes.

%
All in all, the proposed reference-free evaluation protocol
appears a promising 
evaluation method 
to accompany the release of GeNTE  and favour its future utilization as a benchmark for gender-neutral MT.
It proves to be a robust approach, capable of handling the linguistic variability associated with gender neutralization strategies, and overcoming the limitations of the reference-based approaches.



\section{Conclusions}
\label{sec:conclusions}

In this work, we investigated gender-neutral translation as a 
path for inclusive MT.  
To this aim, we focused on English$\rightarrow$Italian,
a pair that is highly representative of the challenges of implementing neutral forms into 
grammatical gender languages. 
As a novel area of inquiry, we started from the fundamentals. First, we conducted a 
survey on the acceptability of gender-neutral translation,
which highlighted the openness of potential MT
end-users, especially in formal communicative situations.
Second, informed by the survey,  we 
built
GeNTE, the first natural benchmark for evaluating gender-neutral translation in MT. Third, we investigated the (un)suitability of existing automatic evaluation protocols to assess gender-neutral translation, and thus proposing an alternative, reference-free solution.
Having taken the first steps toward gender-neutral MT,
our resources and evaluation method are made available to foster and inform
the future
development of more inclusive MT.

\section*{Limitations}
\label{sec:limitations}

Naturally, this work presents some limitations. 
In the paper we took the very first steps to enable evaluation and research on the task of gender-neutral translation for inclusive MT into grammatical gender languages. To do so, we provided data (\S\ref{sec:corpus} \& \S\ref{sec:reference-free}) and modeling ({\S\ref{sec:reference-free}}) for the specific English{$\rightarrow$}Italian language pair.
Thus, except for the GPT prompts written in English, 
the released GeNTE neutral references and the trained classifier
cannot be directly used for other 
target languages.
However, Italian was chosen as a highly representative example of the challenges faced in cross-lingual transfer from English. Accordingly, we believe that our design considerations, the methodology for the creation of GeNTE, as well as the presented 
evaluation protocols
broadly apply to other target grammatical gender languages, too. 

In the experiments, we relied on different closed-source models: Amazon Translate, DeepL and GPT (gpt-3.5-turbo). This has reproducibility consequences, since these models are regularly updated, thus potentially yielding future  results that differ from those reported in this paper. Also, their access via API (paid in US dollars) might not be affordable for all institutions/researchers. 

Finally, due to the inability of current MT 
models to generate gender-neutral translations, 
the output sentences used to test different MT metrics and evaluation protocols (\S\ref{sec:evaluation}) were partially post-edited. Indeed, this solution does not completely reflect standard evaluation practices conducted on fully MT generated output. However, this post-editing process only targeted
gender-related aspects of the output sentences, thus still vastly preserving the MT generation
and offering a controlled, realistic scenario. 
It should be recognized though that, by design, 
we enacted evaluation conditions where the MT models 
\textit{succeeded} in generating the expected (either neutral or gendered) output translation. 
Instead, since the models' outputs did not exhibit cases where MT  \textit{failed} at generating the expected (gendered) translation, we could not test the   
robusteness of our evaluation protocols for such a scenario.



%

\section*{Ethics Statement}
By addressing inclusivity in MT, this work presents an inherent ethical component. It builds from concerns toward the societal impact of widespread translation technologies that reflect and 
propagate discriminatory and exclusionary language. Concretely, by potentially feeding into existing stereotypes, reinforcing male-grounded visibility, and perpetuating the erasure of non-binary gender identities.\footnote{We use non-binary as an umbrella term to encompass all gender identities that do not conform to the Western masculine/feminine binary, and that in particular are not linguistically represented by binary gendered expressions.} Still, our work is not without risks either and thus warrants some ethical considerations. 
In particular, we propose the use of gender-neutralization strategies
that avoid the use of unnecessary gendered terms via the retooling of established forms and grammars. 
These strategies can be considered as a form of Indirect Non-binary Language (INL) \citep{artemistrans}, which are 
intended -- as we do in this paper -- to \textit{equally elicit} all gender identities in language and prevent misgendering by excluding any kind of gender assumption \citep{strengers-2020}.
Instead, Direct Non-binary Language \citep{artemistrans}
-- emerging via grassroots efforts and more predominately in online social medias \citep{lauscher-etal-2022-welcome} --- 
resort to the creation of neologisms, neopronouns or even neomorphology to typically  \textit{enhance} the visibility of non-binary individuals.

In light of the above, several, concurring forms can serve different inclusive language needs \citep{Comandini_2021, knisely2020franccais}. Thus, 
it should be stressed that the neutralization strategies incorporated in our MT work are not prescrivetely intended.
Rather, they are orthogonal to other attempts and non-binary expressions for inclusive language (technologies)  \citep{lauscher2023em, rivas-ginel-2022-all-inclusive}.


\section*{Acknowledgements}

This work is part of the project ``Bias Mitigation and Gender Neutralization Techniques for Automatic Translation'', which is financially supported by an Amazon Research Award AWS AI grant.
Moreover, we acknowledge the support of the PNRR project FAIR - Future AI Research (PE00000013),  under the NRRP MUR program funded by the NextGenerationEU.

\bibliography{custom,anthology}
\bibliographystyle{acl_natbib}

\appendix

\section{Gender-Neutral Translation Survey}
\label{app:questionnaire}

The questionnaire was released online in April 2023. We distributed it via targeted emails and social media
posts, with a request to share with relevant groups.
Participation in this survey was voluntary, uncompensated and anonymous, as no identifying or personal information about the participant was collected.
 Also, participants were free to withdraw at any time without penalty or consequence. 
 An anonymized version of the survey is accessible here: \url{https://forms.gle/YL76UeWbe4NWdCPPA}.

\smallskip
 Note that we did not  target individuals 
 who use MT as a professional tool. Rather, generic stakeholders that might have used MT directly or indirectly (e.g. being offered translations of web pages). They were made aware that results of the questionnaire would have been put to use for research on inclusive MT.
 
\paragraph{Screening questions.}
As the survey required judging English{$\rightarrow$}Italian translations, only participants with high competence of both Italian (C1 or higher) and English (B2 or higher) were eligible to take part in the survey. Accordingly, screening questions aimed to verify such language skills were placed at the beginning of the survey. Of the 101 received responses, 98 were from eligible participants and thus included in our analysis. The screening questions also excluded participants under 18. 

\paragraph{Sociodemographic information.}
To gain sociodemographic information about our participant pools, the survey consisted of a short section asking for background information (e.g., educational level and field), as well as age and self-reported gender information.
Overall, our pool of participants was quite homogeneous in terms of education levels (i.e., the majority of respondents had a master's degree) and age (with 24-35 being most represented age range). 
This was expected, 
because of the channels through which the survey was distributed, but especially in light of the high English competence required to take part in the survey.
In terms of gender, the breakdown in Figure \ref{fig:gender} shows an higher representation of the feminine segment of the population in the survey. Since participation to the survey was voluntary, it might have attracted individuals more interested in the topic. 

We do not consider this homogeneity as a limitation per se. Rather, it allowed us to gauge the opinions of relevant, interested stakeholders, which are mostly affected by discriminatory language. 

\begin{table}[t]
\centering
\footnotesize
\begin{tabular}{|l|r|l}
cisgender woman        & 1  &  \\
female                 & 1  &  \\
woman                  & 55 &  \\
cis male               & 1  &  \\
male                   & 2  &  \\
man                    & 29 &  \\
lad                    & 1  &  \\
non-binary transgender & 1  &  \\
trans man              & 1  &  \\
I don't define myself  & 1  &  \\
--                 & 5  & 
\end{tabular}
\caption{Open responses to the question: \textit{How do you identify?}}
\label{fig:gender}
\end{table}

\paragraph{Linguistic acceptability.}
The pairs of source English sentences aligned with an Italian GT and NT were created by a professional linguist with prior experience on gender-inclusive language. The original source and GT parallel sentences were retrieved from the administrative/legislative domain of EU multilingual documents. The linguist then created the second NT. 

From a methodological perspective, we decided to pair the GT and NT alternatives so to allow for a fixed comparative term. 
Otherwise, different judgments of different NT translations alone would have not allowed for the isolation of gender-related factors from other aspects of the translations that could have influenced participants' 
perception of acceptability. 

\begin{figure}[t]
    \centering
    \includegraphics[scale= 0.45]{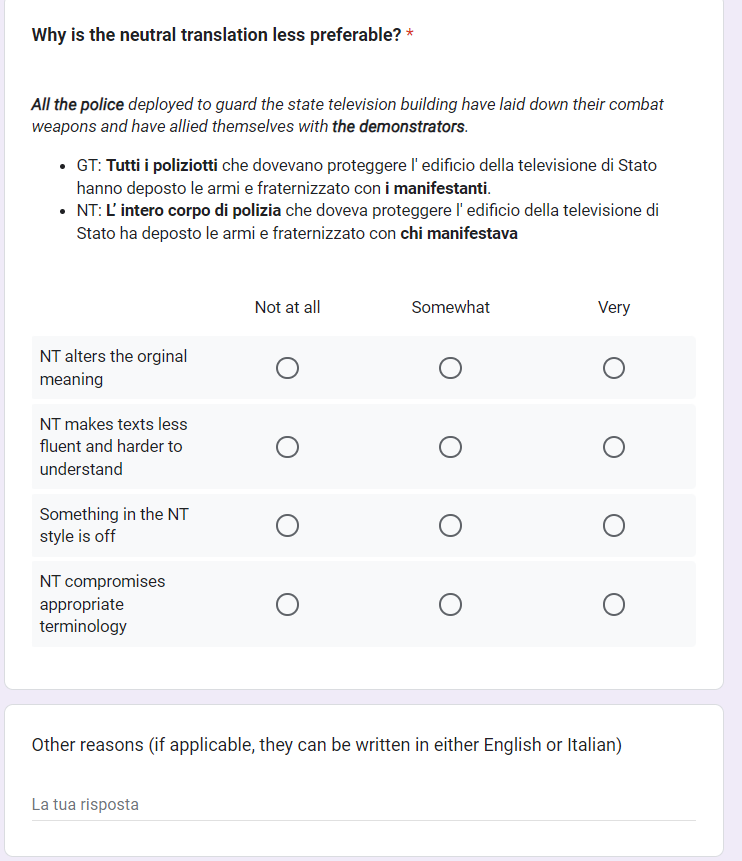}
    \caption{Questionnaire: follow-up questions on linguistic acceptability.}
    \label{fig:follow-up}
\end{figure}

Though not shown in the paper due to space constrains, for each example sentence in the survey the participants were directed to follow up questions, so to motivate their choice and provide 
more insights on the limits of the offered NT (see Figure \ref{fig:follow-up}).
Overall, in this section, a total of 7 example translation were shown. Additionally, for 3 source English sentences, participants were asked to pick a preferred neutral translation from 4 different options. 

\begin{figure}[h]
  \centering
    \includegraphics[scale=0.55]{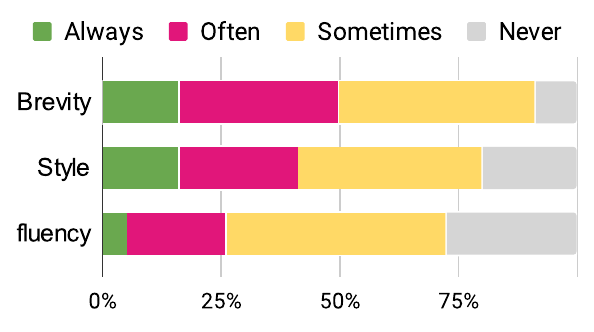} 
  \caption{Willingness to sacrifice different communicative aspects to ensure neutrality.}
  \label{fig:trade-off}
\end{figure}

\paragraph{Use and attitude.}
The last portion of the survey directly 
investigates users' attitude and perception toward gender-neutral language. For instance, in Figure \ref{fig:trade-off}, we attest that participants seem 
inclined to sacrifice brevity 
in favor of neutrality.
Note that, since these questions focus on gender-neutral language -- rather than translation -- they conceptually preceded the section on linguistic acceptability. However, we placed them afterwards, so as 
to avoid influencing participants' responses 
on gender-neutral translations.

\section{GeNTE corpus details}
\label{app:corpus_details}

\subsection{Data editing report}
\label{app:data_editing}


In our data editing process, we performed two types of interventions: \textit{(i)} editing which was functional to the creation and the optimal use of the corpus (motivation described in §\ref{selection}); \textit{(ii)} editing aimed to improve the overall quality of corpus data.

Functional interventions \textit{(i)} include two procedures: (A) the editing of source and reference sentences so as to have them only include referents that require the same type of either neutral or gendered forms (203 entries total); (B) the duplication of gendered entries, in which, then, the gendered words were replaced with equivalents of the opposite gender -- thus we produced 126 masculine entries and 247 feminine ones (373 entries total). Some of the entries underwent both procedures. These procedures were performed on a total of 576 entries.



\begin{table*}[!ht]
    \centering
     \footnotesize
    \begin{tabular}{l|r|c|c|c|c}
        \multicolumn{6}{c}{\textsc{\textbf{common-set-g}}} \\ \hline
        \textbf{↓ REF} & \textbf{CAND →} & \textbf{Reference 1} & \textbf{Reference 2} & \textbf{Reference 3} & \textbf{REF-G} \Tstrut \\ \hline
        \multicolumn{2}{c|}{\textbf{Reference 1}} & - & 73.87 & 76.47 & 73.91 \Tstrut \\ 
        \multicolumn{2}{c|}{\textbf{Reference 2}} & 73.83 & - & 75.21 & 72.07 \\ 
        \multicolumn{2}{c|}{\textbf{Reference 3}} & 76.38 & 75.16 & - & 74.89 \\ 
        \multicolumn{2}{c|}{\textbf{REF-G}} & 73.8 & 71.97 & 74.82 & - \\ \hline 
        \multicolumn{6}{c}{\textsc{\textbf{common-set-n}}} \\ \hline
        \textbf{↓ REF} & \textbf{CAND →} & \textbf{Reference 1} & \textbf{Reference 2} & \textbf{Reference 3} & \textbf{REF-G} \Tstrut \\ \hline
        \multicolumn{2}{c|}{\textbf{Reference 1}} & - & 76.93 & 76.08 & 76.01 \Tstrut \\ 
        \multicolumn{2}{c|}{\textbf{Reference 2}} & 76.95 & - & 76.02 & 73.49 \\ 
        \multicolumn{2}{c|}{\textbf{Reference 3}} & 76.09 & 76 & - & 73.02 \\ 
        \multicolumn{2}{c|}{\textbf{REF-G}} & 75.9 & 73.39 & 72.92 & - \\ \hline
    \end{tabular}
    \caption{BLEU scores representing the linguistic variability in \textsc{common-set}'s references. }
    \label{fig:ling_variability}
\end{table*}

With the second type of interventions \textit{(ii)} we improved the quality of the corpus data.
We did so by correcting translation errors in the original references and removing extra elements in both source and reference sentences. For example, from the source sentence ``EN ) I would like, in particular, to thank Mrs Van den Burg, a Dutch Social Democrat who worked particularly hard on Article 25.'' we removed the segment ``EN )''. We performed these corrections on a total of 89 corpus entries.
Moreover, to improve the variability within the data, we replaced the most frequent noun which entailed a gendered translation, \textit{rapporteur}, with other terms from the institutional/administrative linguistic domain (e.g., \textit{spokesperson}, \textit{delegate, \textit{deputy}}). We performed this operation on 70 corpus entries.

Overall, we edited 314 original source sentences and 393 original reference sentences.


\subsection{Challenges in the creation of gender-neutral references.}
\label{app:ref_creation}

From a qualitative perspective, two main type of challenges were identified in the creation of the neutral references: 

\textbf{Articles}: in 11 instances, the translators produced partial neutralizations, as they overlooked masculine articles. This possibly suggests that, in Italian, articles may be perceived as secondary in expressing gender compared to nouns, adjectives, and verbs, even by native speakers. Regardless, all errors were spotted by the linguist and corrected.

\textbf{Lexical gender}: translators were unable to produce a neutralization for 4 instances of lexically gendered nouns such as `\textit{sorella}’ (sister) and `\textit{figlia}’ (daughter). Such cases all concerned the creation of neutral references for the \textsc{SET-G} – which served the purpose of the contrastive evaluation (Sec. \ref{sec:evaluation}) – but were particularly challenging as they required the use of neutral strategies for unequivocally gendered terms.

Less problematic and systematic difficulties involved specific terms which the translators struggled with, such as `\textit{deputato}’ (deputy). This is possibly due to the fact that some domain-specific terms and their translations, like `deputy-deputato’, are established and rooted in the language to the point where producing a gender-neutral translation is counter-intuitive and challenging. In all cases, the linguist intervened and proposed a solution e.g., `\textit{persona deputata}’ (lit. deputed person).

\subsection{Linguistic diversity in GeNTE's gender-neutral references}
\label{sec:ling_div}

Table~\ref{fig:ling_variability} reports our evaluation of the linguistic variability within the references of the \textsc{common-set}. To perform such evaluation we computed BLEU scores matching 
every reference of each entry with the other two references of that entry. The scores show how there is a noticeable variability, which is attributable to the different neutralization strategies employed by the three translators. On one hand, the rather high BLEU scores indicate that the references are very similar -- as expected, since they share all the original content of the gendered reference, except for the gender-related terms. On the other hand, their distance from a score of 100 BLEU points indicates that they are not perfectly identical.
The variability appears coherent among the sets: the scores of the neutral references evaluated against other neutral references are especially similar in \textsc{common-set-n}, where the highest and the lowest scores differ by less than 1 BLEU point, possibly indicating that the neutralization strategies employed in this set were indeed different, but had very similar impact on the original sentences.


\section{Classifier Training}
\subsection{Generation of Synthetic Training Data}
\label{app:synthetic}

\paragraph{Seed words.}
The generation process began with the creation of  $\sim$200 unique triples of seed words (e.g., \textit{il personale impiegato} [neutral] - \textit{l'impiegato} [masculine] - \textit{l'impiegata} [feminine]).
Half of them 
were sourced from Europarl training data by means of keyword extraction,\footnote{\url{https://pypi.org/project/rake-nltk/}} the other 100 were instead created from scratch. 
We then manually augmented this initial list of triplets by re-generating them with various inflectional morphology, which is relevant to distinguish for the task of neutralization (e.g., make them plural, use indefinite article etc.). Accordingly, we obtained $\sim$800 triplets of seedwords. 

\paragraph{Generation: first round.}

\begin{figure*}[t]
    \centering
    \includegraphics[scale= 0.60]{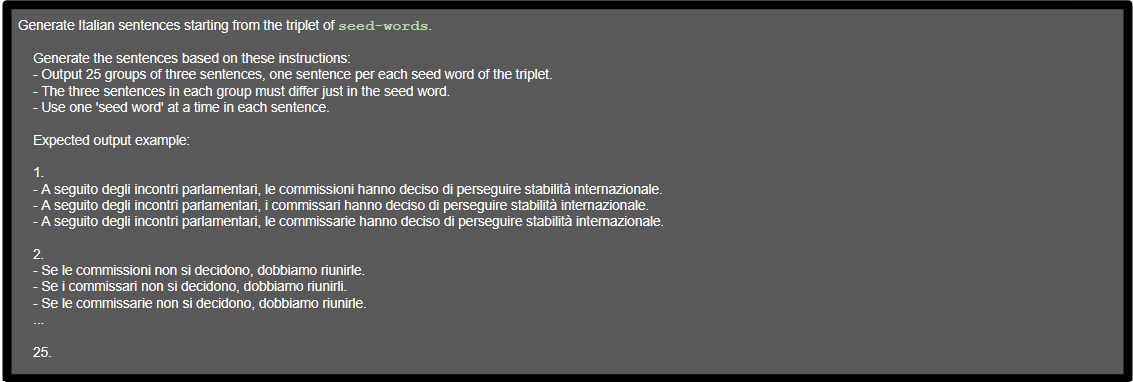}
    \caption{\textbf{Prompt template} for the generation of triplet of sentences from (\textsc{neut/fem/masc}) seed words.}
    \label{fig:First-generation-round}
\end{figure*}

We prompted the \texttt{gpt-3.5-turbo} model from the GPT LLM family\footnote{\url{https://platform.openai.com/docs/models/gpt-3-5}} to generate triplets of sentences given the triplets of seed words. 
We used a few-shot approach with given examples of the task to be performed (see Figure \ref{fig:First-generation-round}.).  
We access the model via OpenAI paid API and setting a temperature of 0.5.\footnote{\url{https://platform.openai.com/docs/api-reference}} 
In total, approximately 60,000 sentences were generated. A random sample of 100 sentences was manually inspected, revealing that noise was very low (10\%), but that the sentences exhibited a simple structure, consistently placing the subject at the beginning.

\begin{figure*}[t]
    \centering
    \includegraphics[scale= 0.60]{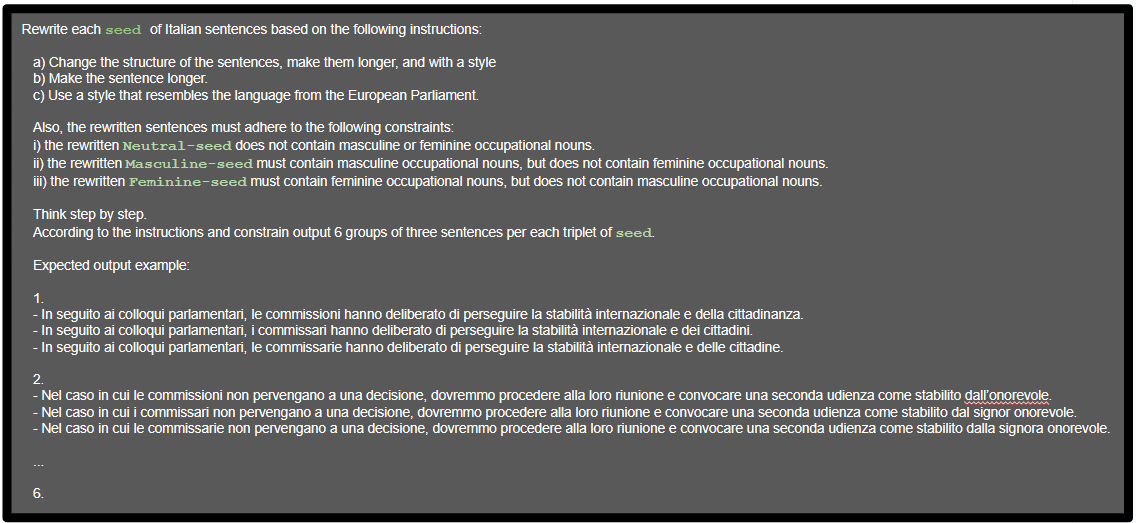}
    \caption{\textbf{Prompt Template} for the rewriting of triplet of (\textsc{neut/fem/masc}) seed sentences.}
    \label{fig:second-generation-round.}
\end{figure*}

\paragraph{Generation: second round.}

To enhance the quality and textual context of the generated sentences, 
a second round of generation was performed using a lower temperature of 0.3 (see Figure \ref{fig:second-generation-round.}).
Each triplet of sentences was rewritten multiple times in different forms. This process resulted in the generation of approximately 320,000 sentences,
which had
a higher occurrence of incorrect alternatives for the seedwords,
estimated to be around 40\% based on the inspection of 100 randomly selected sentences. 
The final synthetic corpus consists of approximately 380,000 sentences, 
featuring varied sentence structures. Specifically, one-third of the sentences contain a masculine seedword, another third contain a feminine seedword, and the remaining third contain a neutral seedword.

Overall,  a cost of \$13.12 USD was estimated.

\subsection{Training Setup}
\label{app:training_setup}

We trained the parameters of both the linear layer and UmBERTo on the classification task for 2 epoch, with learning rate of 5e-5, batch size of 64 and maximum sequence length of 64, on a p3.2xlarge instance on AWS
(featuring one NVIDIA V100 GPU).
The code for finetuning relies on Huggingface transformers library \citep{wolf-etal-2020-transformers}.
%

\end{document}